\documentclass[10pt,twocolumn,letterpaper]{article}

\usepackage[pagenumbers]{wacv} 
\usepackage[strings]{underscore}
\definecolor{wacvblue}{rgb}{0.21,0.49,0.74}
\usepackage[pagebackref,breaklinks,colorlinks,allcolors=wacvblue]{hyperref}
\usepackage{tabularx}
\usepackage{multirow}
\usepackage{amssymb}
\usepackage{pifont}
\newcommand{\cmark}{\ding{51}}
\newcommand{\xmark}{\ding{55}}

\title{SENSE-VAD: Sentient and Semantic Video Anomaly Detection for Autonomous Driving}

\author{
Nghia T. Nguyen$^{*}$\\
University of South Florida\\
{\tt\small nghianguyen@usf.edu}
\and
Lokman Bekit$^{*}$\\
University of South Florida\\
{\tt\small lbekit@usf.edu}
\and
Yasin Yilmaz\\
University of South Florida\\
{\tt\small yasiny@usf.edu}
}

\begin{document}
\maketitle
\renewcommand{\thefootnote}{}
\footnotetext{$^{*}$Equal contribution.}
\renewcommand{\thefootnote}{\arabic{footnote}}

\begin{abstract}
Autonomous vehicles (AVs) must navigate not only motion-based hazards but also socially complex situations whose danger is constituted by inter-agent relationships rather than movement statistics alone. A child running away from a guardian, a person being carried by another, or a pursuer chasing a pedestrian across a sidewalk are all anomalous in social context, yet none produces an obvious motion signal that current anomaly detectors are equipped to flag. We introduce \textit{SENSE-VAD}, the first synthetic video anomaly detection benchmark for autonomous driving explicitly designed around socially complex anomalies. Using the CARLA simulator and Unreal Engine (UE), we generate distinct anomaly scenarios across multiple categories: individual behaviors, group behaviors, person--object interactions, cyclist interactions, vehicle \& agent, each annotated with per-frame binary labels. A key design principle is the separation of social anomaly from motion-based or appearance-based anomaly: many scenarios involve motion of objects that appears unremarkable in isolation but is anomalous in relational context. We additionally provide real-world normal and anomalous videos as a sim-to-real transfer probe. We evaluate state-of-the-art video anomaly detection baselines and demonstrate that socially complex anomalies constitute a distinct and currently unsolved challenge. Our dataset, annotations, and generation code are publicly available.\footnote{\url{https://zenodo.org/records/20955310}}

\end{abstract}

\vspace{-5mm}
\section{Introduction}
\label{sec:intro}
\vspace{-2mm}
Autonomous vehicles (AVs) have made remarkable progress in detecting and responding to conventional traffic hazards: lane departures, stationary obstacles, and kinematically irregular trajectories. Yet a class of
safety-critical scenarios remains largely unaddressed, anomalies whose danger is not encoded in motion statistics, but in the \emph{social context}
surrounding that motion. A child, whose actions are naturally less predictable than adults, running along a sidewalk presents a very different risk profile depending on whether a parent is nearby, whether the
child is moving toward or away from the road, and whether a pursuing adult is visible in the scene. A group of people sprinting across a street may signal a
police chase, a medical emergency, or a spontaneous celebration, each demanding a different vehicular response. Current AV perception systems,
optimized for appearance- and trajectory-based signals, are structurally blind to these distinctions.

The June 2026 Waymo recall illustrates a gap that incremental software patches cannot close. After more than a dozen incidents in which its vehicles failed to recognize ramp-closure signs and drove into active freeway construction zones, the company recalled nearly 3{,}900 robotaxis, its second recall in just over a month and one preceded by failures ranging from entering a flooded lane to a collision with a child near an elementary school~\cite{waymo_recall_reuters}. The pattern is telling: each recall resolves a specific, previously unseen scenario, yet the long tail continually produces new ones, and no amount of per-case engineering can enumerate every situation an AV will meet in the open world. What is needed is not another patch but a \emph{warning system} that recognizes when the current scene is anomalous and falls outside the model's competence, and that can hand off to a human or a higher-level reasoning agent before the vehicle acts. Building such a system requires data that captures the socially structured yet motion-statistically plausible events these failures exemplify, precisely the gap SENSE-VAD is designed to fill.

\begin{figure}[t]
  \centering
  \includegraphics[width=\columnwidth]{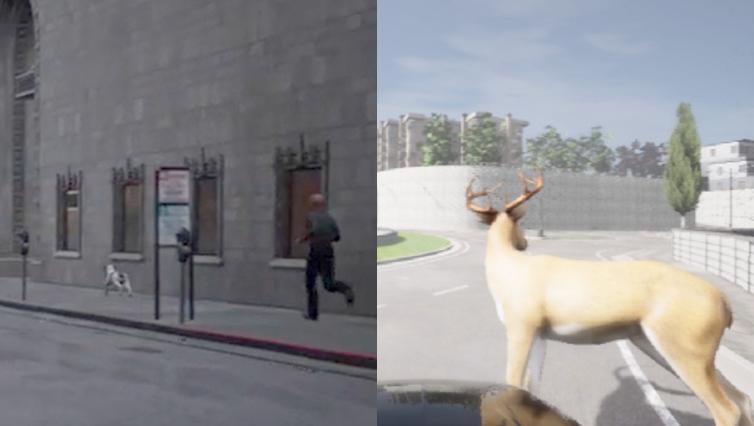}
\caption{Qualitative comparison between SENSE-VAD (left) and AnoVox~\cite{bogdoll2024anovox} (right).
SENSE-VAD captures socially-defined anomalies (\textit{e.g.}, a person chasing their dog and the path is not toward the road), whereas AnoVox focuses on single-object anomalies
(\textit{e.g.}, an animal on the roadway).}
  \label{fig:qualitative_comparison}
\end{figure}

Existing video anomaly detection (VAD) datasets for autonomous driving, including DADA~\cite{fang2021dada}, DoTA~\cite{yao2022dota}, and CADP~\cite{shah2018cadp}, focus primarily on traffic accidents, vehicle-centric and single-object anomalies.
Pedestrian-focused and top-view trajectory
datasets~\cite{liu2025pfsd,mbuya2025graph} address crowd behavior but lack the egocentric AV
perspective, fine-grained social-relational annotations, and the diversity of
subtle behavioral cues this problem requires. Crucially, none of these
resources defines or systematically annotates anomalies whose classification
requires reasoning about inter-agent social relationships (Fig. \ref{fig:qualitative_comparison}).

To address this gap, we introduce \textbf{SENSE-VAD}, the first synthetic video anomaly detection dataset for autonomous driving designed explicitly around \emph{socially complex anomalies}. The dataset is generated with the CARLA simulator~\cite{dosovitskiy2017carla} across two engine generations: CARLA 0.9.12 and 0.9.15 on Unreal Engine~4 (UE4), and CARLA 0.10.0 on Unreal Engine~5 (UE5), with the version chosen per scenario according to its animation and agent-control requirements. On UE5 we additionally build scenes from the City Sample project to diversify the visual environment beyond CARLA's native
maps. SENSE-VAD covers \textbf{15} anomaly classes across six conceptual categories, each is represented by multiple
video clips with per-frame anomaly labels, yielding \textbf{108} annotated clips in total (Table~\ref{tab:ds_stat}).

A key design principle of our dataset is the separation of \emph{social anomaly}
from \emph{motion-statistics anomaly}. Many of our scenarios, such as a child running
away from a parent, a bag left unattended on a pavement, a person being carried
by another, involve motion that appears normal in isolation but is anomalous
in its social context. We validate that this distinction poses a genuine
challenge to current methods by benchmarking 11 state-of-the-art VAD models on our dataset, showing that methods which perform competitively on kinematic anomalies suffer measurable degradation on socially complex scenarios.

To support generalization beyond synthetic data, we additionally include a set
of real-world normal and anomalous video clips collected from naturalistic
settings (e.g., persons acting erratically in public spaces). These serve as a
qualitative domain-transfer probe, helping ensure that models performing well on
our benchmark have learned genuine human behavioral patterns rather than
simulator-specific visual artifacts.
Our contributions are as follows:
\begin{itemize}
    \item We present the first taxonomy of \emph{socially complex anomalies} for
    autonomous driving, comprising \textbf{15} classes across six categories,
    each defined to distinguish social anomaly from motion-statistics anomaly.

    \item We introduce \textbf{SENSE-VAD}, a CARLA-generated benchmark with per-frame labels, built across three CARLA versions to meet the animation and agent-control requirements of each scenario type, and complemented by real-world clips, both normal and anomalous, that probe sim-to-real transfer. We also release the parameterized generation scripts for authoring new scenarios.

    \item We benchmark \textbf{11} state-of-the-art VAD methods and show that
    socially complex anomalies remain a distinct and currently unsolved
    challenge for AV perception.
\end{itemize}
\vspace{-2mm}
\section{Related Work}
\label{sec:related}
\vspace{-1mm}
\textbf{Video Anomaly Detection:}
Video anomaly detection (VAD) has matured across three supervision regimes,
evaluated largely on a common set of benchmarks, UCSD Ped~\cite{li2013anomaly},
CUHK Avenue~\cite{lu2013abnormal}, ShanghaiTech~\cite{liu2018ano_pred}, Street
Scene~\cite{ramachandra2020street}, UCF-Crime~\cite{sultani2018real},
XD-Violence~\cite{wu2020not}, and IITB-Corridor ~\cite{rodrigues2020multi}. Semi-supervised methods train only on normal video and flag
deviations via reconstruction or future-frame prediction
error~\cite{hasan2016learning,liu2018future,gong2019memorizing,georgescu2021anomaly};
weakly-supervised methods exploit video-level anomaly labels under multiple-instance
learning~\cite{sultani2018real,chen2023mgfn,zhou2024batchnorm,karim2024real};
and unsupervised methods derive pseudo-labels from feature statistics
alone~\cite{shi2024learning,al2024collaborative}. The recent training-free approach
replaces learned normality with the scene understanding of foundation models,
scoring LLM-generated captions, querying multimodal LLMs, or inducing
scene-specific normality
rules~\cite{zanella2024harnessing,yang2024follow,ye2025vera,wu2024vadclip,li2026vadtree,mumcu2025leveraging}.
We benchmark representatives of all four regimes (Section~\ref{sec:baselines}).

The closest work to ours, ComplexVAD~\cite{mumcu2025complexvad}, targets \emph{interaction}-based anomalies, where
individual agents behave normally but their relational dynamics are anomalous. This is a surveillance dataset  recorded at a university
crosswalk in which anomalies arise from inter-agent
interactions. This validates relational anomalies as a
distinct problem, but the setting is limited to a \emph{fixed surveillance camera} with
no ego-centric perspective, no driving-relevant social taxonomy, and no
sim-to-real probe. More broadly, existing VAD datasets are built around a fixed
observer. Even datasets aggregated from heterogeneous web sources, such as
XD-Violence~\cite{wu2020not}, contain only incidental in-vehicle footage and are
not designed for the onboard viewpoint, and they define anomalies through motion
statistics, reconstruction error, or crime-category labels rather than social
context. None targets the egocentric AV perspective or the socially contextual
anomalies that govern pedestrian safety.
 
\textbf{Anomaly Datasets for Autonomous Driving:}
A separate body of work builds anomaly datasets for the AV domain, but uniformly
defines the anomaly through vehicle kinematics or accident outcomes. Dashcam
accident datasets provide ego-centric clips with temporal or spatial accident
annotations~\cite{chan2016anticipating,yao2019unsupervised,fang2021dada,shah2018cadp,singh2020anomalous}, of which DoTA is the most
comprehensive ($4{,}677$ clips, 18 categories, spatio-temporal
boxes)~\cite{yao2022dota} and DRAMA additionally couples risk to textual
descriptions~\cite{malla2023drama}. Out-of-distribution segmentation benchmarks instead
define the anomaly as a pixel of an unseen
class~\cite{pinggera2016lost,blum2021fishyscapes,singh2020anomalous,li2022coda}, a purely visual-semantic
notion with no relational component. Simulation-based benchmarks scale anomaly
diversity in CARLA or game engines~\cite{gong2024sdac,wang2024deepaccident,kim2019crash}, with
AnoVox being the largest synthetic AV anomaly set~\cite{bogdoll2024anovox}.

In these existing datasets, the anomaly is vehicle-centric or post-hoc: a
collision, a corner-case object, a kinematic outlier. None systematically captures the socially complex, inter-agent events, that are critical for pedestrian safety. More details about these differences are shown in Table~\ref{tab:dataset_comparison}.

\textbf{Language Models and Long-tail Generalization in AV:}
Recent work integrates VLMs and LLMs into the AV stack for explainable perception and planning~\cite{drivelm2024,chen2023drivingllmsfusingobjectlevel,chen2024asynchronouslargelanguagemodel,wang2025omnidriveholisticvisionlanguagedataset,
ishida2024langpropcodeoptimizationframework,nie2024reason2driveinterpretablechainbasedreasoning}. We adopt the structured question types of DriveLM~\cite{drivelm2024} in our probing study (Section~\ref{sec:probe}). Two findings motivate that study. First, even LLM-augmented planners fail on rare
long-tail scenarios: on interPlan~\cite{hallgarten2024can}, augmenting nuPlan with
construction zones, accidents, and jaywalkers collapse state-of-the-art
planners that otherwise score near-perfectly. Second, \cite{xie2025vlmsreadyautonomousdriving} shows via DriveBench that VLMs frequently produce visually ungrounded driving responses, performing comparably with or without visual input because outputs are dominated by language priors. The long-tail scenarios these benchmarks
omit are precisely the socially complex pedestrian anomalies we target, and the ungroundedness they expose is exactly the failure SENSE-VAD is built to measure and, through socially annotated training data, to address.

\begin{table}[t]
\centering
\resizebox{\columnwidth}{!}{%
\begin{tabular}{lrccccc}
\toprule
\textbf{Dataset} & \textbf{\#Frames} & \textbf{Ego} & \textbf{Temp.} &
\textbf{Social} & \textbf{Sim.} & \textbf{Spatial} \\
\midrule
ComplexVAD~\cite{mumcu2025complexvad}         & $\sim$3.68M  & \xmark & \cmark & \cmark & \xmark & \cmark \\
DoTA~\cite{yao2022dota}                     & ~297K           & \cmark & \cmark & \xmark & \xmark & \cmark \\
CODA~\cite{li2022coda}                     & 1{,}500      & \cmark & \xmark & \xmark & \xmark & \cmark \\
DeepAccident~\cite{wang2024deepaccident}     & 57{,}000     & \cmark & \cmark & \xmark & \cmark & \cmark \\
AnoVox~\cite{bogdoll2024anovox}                 & 245{,}600    & \cmark & \cmark & \xmark & \cmark & \cmark \\
\midrule
\textbf{SENSE-VAD (ours)}            & \textbf{540{,}888} & \cmark & \cmark & \cmark & \cmark & \cmark \\
\bottomrule
\end{tabular}%
}
\caption{Representative anomaly-detection datasets. ``Social'' = anomalies
defined by inter-agent relationships rather than appearance, motion, or
accident outcome; ``Spatial'' = provides spatial (box/mask) annotation.
SENSE-VAD is the only benchmark that is egocentric, per-frame temporal,
social, and simulated at once. \cmark/\xmark = has/lacks.}
\label{tab:dataset_comparison}
\end{table}
\vspace{-2mm}
\section{Dataset and Preprocessing}
\label{sec:dataset}
\vspace{-1mm}

\subsection{Design Principles}
\textbf{SENSE-VAD} is built around a single guiding principle: anomaly labels
should reflect \emph{social context}, not isolated motion. The danger of a
scenario is encoded in the relationship between agents. A child running away
from a parent is anomalous not because of the child's speed but because of the
relational configuration: an unsupervised child, a diverging direction of
movement, and proximity to a road. This principle has two practical
consequences. First, we define a formal taxonomy of 15 socially complex anomaly
classes (Section~\ref{subsec:taxonomy}). Second, we combine synthetic anomaly
clips with diverse normal footage drawn from multiple sources, so that models
cannot succeed by fitting to simulator-specific visual patterns, and we further
record real-world clips, both normal and anomalous, to test transfer beyond
simulation (Section~\ref{subsec:normal}).
\subsection{Anomaly Taxonomy}
\label{subsec:taxonomy}
We organize our anomaly scenarios into 15 classes across six conceptual
categories. This consolidation is principled: scenarios sharing the same social
cue and AV-response requirement are merged into a single class, while scenarios
with qualitatively distinct relational structures are kept separate.
Table~\ref{tab:taxonomy} provides the full taxonomy. Severity ratings reflect
the urgency of an appropriate AV response: high-severity classes demand
immediate deceleration or trajectory modification, whereas medium and low
classes warrant heightened attention without necessarily requiring immediate
action.

\begin{table*}[ht]
\centering
\small
\setlength{\tabcolsep}{5pt}
\renewcommand{\arraystretch}{1.2}
\begin{tabularx}{\textwidth}{|l|l|X|c|}
\hline
\textbf{Category} & \textbf{Class} & \textbf{Social Cue} & \textbf{Sev.} \\
\hline
\multirow{3}{*}{Child Safety}
  & Child Running from Guardian
    & Unsupervised child diverging from adult caregiver
    & High \\ \cline{2-4}
  & Kidnapping
    & Adult forcibly removing child against their will
    & High \\ \cline{2-4}
  & Kid Occlusion
    & Child temporarily hidden from AV view, causing safety uncertainty
    & Med  \\
\hline
\multirow{2}{*}{Group Behavior}
  & Police Chase
    & Directed pursuit dynamic across or near traffic
    & High \\ \cline{2-4}
  & Crowd Chaos
    & Unpredictable collective movement spilling into the carriageway
    & High \\
\hline
\multirow{2}{*}{Individual Behavior}
  & Abnormal Person Behavior
    & Erratic movement or loss of motor control near traffic
    & Med  \\ \cline{2-4}
  & Person Being Carried
    & One agent physically transporting another, signalling injury or coercion
    & High \\
\hline
\multirow{3}{*}{Person--Object Interaction}
  & Bag Theft
    & Sudden forced removal of property followed by flight
    & High \\ \cline{2-4}
  & Unattended Bag
    & Object abandoned near road creating suspicion or obstruction hazard
    & Low  \\ \cline{2-4}
  & Object Drop
    & Fallen item creating sudden obstruction in or near the carriageway
    & Med  \\
\hline
\multirow{3}{*}{Cyclist Interaction}
  & Cyclist Hazard
    & Loss of control or erratic trajectory placing cyclist near traffic
    & Med  \\ \cline{2-4}
  & Cyclist--Bag Abandonment
    & Bag deposited on or near road by a cyclist, creating lateral obstruction
    & Low  \\ \cline{2-4}
  & Cyclist--Vehicle Collision
    & Direct impact between cyclist and motor vehicle
    & High \\
\hline
\multirow{2}{*}{Vehicle \& Agent}
  & Vehicle--Pedestrian Collision
    & Pedestrian struck by vehicle with secondary hazard cascade
    & High \\ \cline{2-4}
  & Dog Leaving Owner
    & Animal may enter road unsupervised, analogous to unsupervised child
    & High \\
\hline
\end{tabularx}
\caption{SENSE-VAD anomaly taxonomy across 15 classes and six categories.
Severity reflects urgency of AV response: \textit{High} demands immediate
deceleration or trajectory modification; \textit{Med} warrants heightened
attention; \textit{Low} requires awareness without immediate action.}
\label{tab:taxonomy}
\end{table*}

The taxonomy groups classes by the cue an AV must reason about rather than by
surface appearance. \textit{Child Safety} is organized around relational
dynamics: whether motion is child-initiated, adult-coerced, or simply
occluded. \textit{Group Behavior} separates emergent collective motion
(Crowd Chaos) from structured pursuit (Police Chase).
\textit{Individual Behavior} and \textit{Object Interaction} isolate
single-agent anomalies and property- or obstruction-related events,
respectively. \textit{Cyclist Interaction} consolidates hazards sharing a
single cue; a cyclist whose trajectory or cargo becomes unreliable.
Finally, \textit{Vehicle \& Agent} includes collision and Dog Leaving Owner since an unaccompanied animal in the
road poses a hazard analogous to an unsupervised child.
\subsection{Simulation Platform}
\label{subsec:platform}
Our anomaly clips are generated using three versions of the CARLA open
autonomous-driving simulator~\cite{dosovitskiy2017carla}, with version selection determined by
the technical requirements of each class. Using multiple engine builds is a
deliberate choice: no single CARLA release supports the full range of
multi-agent AI, skeletal control, and animal locomotion our taxonomy demands,
so we pair each scenario type with the build that renders it most faithfully.
\subsubsection{CARLA 0.9.12} 
This version is used for scenarios requiring complex multi-agent
pedestrian AI: Crowd Chaos, Police Chase, Child Running from Guardian, and Kidnapping. Its navigation control enables realistic emergent group behavior and pursuit dynamics. Its primary limitation is that the \texttt{get\_bones()}
method is non-functional, preventing direct object attachment to skeletal
bones. We address this for carry scenes by implementing a sinusoidal arm-motion
function that replicates natural carry dynamics without bone attachment.
\subsubsection{CARLA 0.9.15} This version is used for Bag Theft, Unattended Bag, Cyclist Hazard,
Vehicle--Pedestrian Collision, and Abnormal Person Behavior. The functional
\texttt{get\_bones()} API in 0.9.15 enables natural object attachment to arm
bones, which is essential for the bag-related classes.
\subsubsection{CARLA 0.10.0 on Unreal Engine~5} This version is used for the Dog Leaving Owner
class, where its animal-skeleton packages and enhanced bone control produce
more realistic animal locomotion than earlier builds support natively.
\subsubsection{Other experimented platforms} 
We evaluated alternative platforms including GTA\,V, CarSim, and Euro Truck
Simulator (ETS), as well as text-to-video (TTV) generative models such as
Sora~\cite{sora} and Veo\,3~\cite{veo}. GTA\,V was rejected due to its
closed-source engine and simplified physics. CarSim and ETS lack sufficient
pedestrian-interaction control. TTV models ~\cite{sun2025terasim} produce non-deterministic physics
violations in complex multi-agent scenes and prohibit child characters by
content policy, a critical limitation given that several of our
classes directly involve children or social coercion. We also evaluated Cosmos
Transfer~\cite{nvidia2025cosmos} as a photorealism post-processing step but found that it
compromised the physical-interaction fidelity essential for anomaly
interpretation.

\subsection{Scene Construction}
\label{subsec:scenes}
\begin{figure*}[t]
  \centering
  \includegraphics[width=0.9\textwidth]{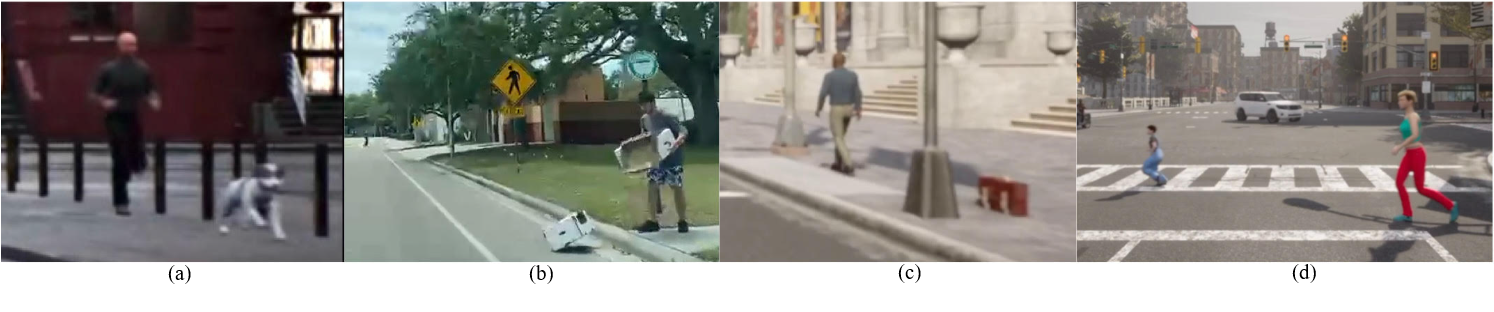}
    \vspace{-3mm}
    \captionsetup{justification=centering}
  \caption{Representative frames from the complex anomaly videos in SENSE-VAD.}
  \label{fig:anomalies}
    \vspace{-4mm}
\end{figure*}

Each anomaly clip is constructed as a scripted CARLA or UE5 episode with controlled agent placement, camera configuration, weather, and timing. We adopt an egocentric dashcam perspective recording at 25--60 frames per second at a resolution of \textbf{$800 \times 600$}. Camera parameters are held constant across all scenarios to ensure comparability across classes.To prevent models from learning class-specific background cues, we vary the CARLA town map, time of day, and weather conditions across clips within each class. Each of the 15 classes is represented by 1--5 clips, yielding 65 anomaly clips in the current release (21 training, 44 test).
Additionally, the real-world clips in the test set are captured from a moving car or motorcycle to match the egocentric, vehicle-mounted viewpoint of our synthetic data, and depict adult-only social anomalies that can be staged safely and with consent, e.g., a person dropping boxes onto the road or a pedestrian stumbling near traffic.

Figure~\ref{fig:anomalies} shows representative anomalies from the dataset: (a) a dog leaving its owner, (b) an object dropped near the carriageway, (c) a bag abandoned at the roadside, and (d) a child running away from a guardian. Each is defined by an inter-agent social relationship rather than a motion outlier. Panels (a), (c), and (d) are CARLA-simulated scenes, while (b) is a real-world clip from the test set.

\subsection{Annotation}
\label{subsec:annotation}

Per-frame binary anomaly labels are assigned according to the following
protocol. The \emph{anomaly onset frame} is the first frame in which the
socially anomalous condition is unambiguously established: for example, the
frame at which a child's trajectory diverges from the guardian, the frame at
which a bag is placed and the depositor begins to walk away, or the frame of
first vehicle--pedestrian contact. All frames from onset to clip end are
labelled anomalous; pre-onset frames are labelled normal. This scheme is
conservative: ambiguous frames are labelled normal to avoid inflating anomaly
recall.

\subsection{Normal Video Sources}
\label{subsec:normal}

A robust anomaly-detection benchmark requires a large and diverse pool of normal
footage. Relying solely on our own CARLA recordings would risk models separating
anomaly from normality on the basis of simulator-specific visual cues rather
than genuine behavioural signals. We therefore construct the normal split from
three complementary sources.

\subsubsection{CARLA synthetic normal clips} For each anomaly class we generate a
matched normal clip in the same scene configuration in which the anomalous
event does not occur. Critically, all \emph{targeted} normal counterparts, the behavioral near-misses that a model could otherwise exploit, such as a child
walking calmly beside a guardian or a dog kept on a leash beside its owner, are produced in simulation, so that each anomaly has a behaviorally matched normal twin that differs only in the social cue, not in appearance or scene. This paired structure supports direct per-class anomaly-localization evaluation.

\subsubsection{DAVID dataset} To enlarge and diversify the normal distribution beyond
our paired recordings, we incorporate the CARLA Densely Annotated Driving
Dataset, which we abbreviate DAVID~\cite{david}. It consists of 28 video
sequences of urban driving recorded in the CARLA simulator, totalling 10{,}767
frames across sunny, rainy, and cloudy conditions, and covering regular driving,
traffic jams, and traffic-light interactions.

\subsubsection{Real-world clips} To probe sim-to-real transfer, we additionally record
real-world clips containing \emph{both} normal driving and staged anomalous
behavior (e.g.\ a person acting erratically near a roadway). These real
anomalies are included to prevent a model from scoring well on synthetic data by learning CARLA's rendering and not transfering to real data. By evaluating whether models trained or validated on our synthetic benchmark also flag genuine real-world anomalies, we provide direct evidence that performance reflects learned social-behavioral structure rather than simulator-specific artifacts.

\subsection{Dataset Statistics}
\label{sec:ds_stat}

Table~\ref{tab:ds_overview} summarizes the composition of SENSE-VAD, and
Table~\ref{tab:ds_stat} reports its train/test partition. The split is
deliberately weighted toward evaluation: the test partition contains more
anomaly clips and frames than the training partition. It further introduces
three anomaly classes (kid occlusion, cyclist vehicle collision,
and kidnapping) absent from training, yielding an open-set evaluation in
which methods must generalize to anomaly types unseen during training. At roughly half a million frames, the benchmark is large enough to exercise every supervision regime while remaining tractable for the computationally expensive vision–language baselines of 
Section~\ref{sec:baselines}.

\begin{table}[ht]
\centering
\resizebox{\columnwidth}{!}{%
\begin{tabular}{ll}
\toprule
\textbf{Property} & \textbf{Value} \\
\midrule
Anomaly classes              & 15 \\
Categories                   & 6 \\
Clips per class              & 1--5 \\
Total anomaly clips          & 65 \\
Normal clips (CARLA)         & 13 \\
Normal sequences (DAVID)     & 28 sequences / 10{,}767 frames \\
Real-world Anomalous clips             & 4  \\
Real-world Normal clips             & 2  \\
Simulator software           & CARLA (UE4 \& UE5) \\
Annotation type              & Per-frame binary labels \\
Camera perspective           & Egocentric (AV dashcam) \\
Resolution                   & $800 \times 600$ \\
Frame rate                   & 25--60\,FPS \\
\bottomrule
\end{tabular}%
}
\caption{SENSE-VAD dataset overview.}
\label{tab:ds_overview}
\end{table}

\begin{table}[ht]
\centering
\begin{tabular}{lrrr}
\toprule
                & Train   & Test    & Total            \\
\midrule
Anomaly clips   & 21      & 44      & \textbf{65}      \\
Normal clips    & 27      & 16      & \textbf{43}      \\
Total clips     & 48      & 60      & \textbf{108}     \\
Total frames    & 262{,}534 & 278{,}354 & \textbf{540{,}888} \\
Avg.\ frames/clip & 5{,}469 & 4{,}639 & \textbf{5{,}008}  \\
Duration (min)  & 80.0    & 91.2    & \textbf{171.2}   \\
\bottomrule
\end{tabular}
\caption{SENSE-VAD dataset statistics.}
\label{tab:ds_stat}
  \vspace{-4mm}
\end{table}
\section{Experimental Setup}
\label{sec:exp_setup}
\subsection{Evaluation Protocol}
\label{sec:eval_protocol}
All experiments are conducted on a single NVIDIA RTX 5090 (32 GB VRAM), with VadCLIP and VERA trained in mixed-precision BF16 and all remaining methods applied at inference time without parameter updates. We evaluate all baselines using frame-level Area Under the ROC Curve (AUC) as the primary metric, consistent with established VAD benchmarks. We assign frame-level ground-truth labels following the protocol of Section ~\ref{subsec:annotation}. For clips with multiple anomaly intervals (e.g \texttt{police_chase_bag_drop}; \texttt{drop_bag_child_running}), the onset/offset pairs are unioned into a single label vector, and the per-clip intervals are listed in the supplementary material. Each method's frame-level scores are then linearly interpolated to the clip's frame count before AUC computation.

\subsection{Baselines}
\label{sec:baselines}
We select baselines by a fixed protocol rather than by convenience. Because SENSE-VAD provides splits for every supervision regime, we cover all four VAD paradigms recognized in the literature: semi-supervised (normal-only), weakly-supervised (video-level MIL), unsupervised (no labels), and training-free foundation-model reasoning. Within each regime we sample distinct algorithmic families, spanning reconstruction/prediction, feature-distribution statistics, MIL/metric learning, and LLM/VLM reasoning, so that any drop on social anomalies cannot be attributed to a single algorithmic blind spot. Each method is either the canonical reference for its paradigm (e.g., Future Frame Prediction (FFP)) or a 2023--2025 state-of-the-art. We deliberately over-sample the semantic VLM/LLM family (LAVAD, AnomalyRuler, VADTree, and the VLM-adapted VadCLIP and VERA), since these explicitly claim relational scene understanding and are the strongest a priori candidates for detecting social anomalies; if even they fail, the gap is established conservatively. For reproducibility, we use a shared Video Swin-B extractor wherever an original extractor was unavailable, controlling for backbone confounds, and defer full hyperparameters and the LAVAD scoring prompt to the supplementary.
Details of benchmarked methods are provided in the Appendix.

\section{Benchmark Results}
\label{sec:results}

\subsection{Benchmark Result Summary}


Table~\ref{tab:main_results} reports frame-level AUC for all eleven baselines, grouped by supervision paradigm. No method exceeds 57.2\% AUC, and the 19.8-point spread confirms meaningful discrimination across approaches. Three patterns emerge: kinematic methods are actively inverted (REWARD 37.38\%, BN-WVAD 37.86\%, both below chance), the standard supervision hierarchy is reversed so that unsupervised methods top the leaderboard, and VLM-based methods cluster near chance. Because the real-world clips are part of the combined test set, every reported AUC already reflects a mix of synthetic and real frames, so each method is exercised on genuine footage as well as simulation. Together with the probing results in Figure~\ref{fig:probe_tprfpr}, where Qwen3-VL-4B-Instruct fails to acknowledge anomalous frames, these findings indicate that progress on SENSE-VAD requires explicit reasoning about inter-agent social relationships rather than refinements to the per-agent kinematic or appearance signals on which existing methods rely.

\begin{table}[t]
\centering
\small
\setlength{\tabcolsep}{5pt}
\renewcommand{\arraystretch}{1.1}
\begin{tabular}{lccc}
\toprule
\textbf{Method} & \textbf{AUC (\%)} & \textbf{Real-time} & \textbf{Explain.} \\
\midrule
\multicolumn{4}{@{}l}{\textit{Semi-supervised --- normal-only training}} \\
FFP~\cite{liu2018future}            & 44.56 & \xmark & \xmark \\
\midrule
\multicolumn{4}{@{}l}{\textit{Weakly-supervised --- video-level MIL}} \\
VadCLIP~\cite{wu2024vadclip}        & \underline{50.80} & \xmark & \xmark \\
VERA~\cite{ye2025vera}              & 50.53 & \xmark & \cmark \\
REWARD~\cite{karim2024real}         & 37.38 & \cmark & \xmark \\
BN-WVAD~\cite{zhou2024batchnorm}       & 37.86 & \cmark & \xmark \\
\midrule
\multicolumn{4}{@{}l}{\textit{Unsupervised --- no labels}} \\
NP-VAD~\cite{shi2024learning}       & \underline{\textbf{57.15}} & \cmark & \xmark \\
CLAP~\cite{al2024collaborative}     & 56.12 & \cmark & \xmark \\
\midrule
\multicolumn{4}{@{}l}{\textit{Training-free --- foundation-model reasoning}} \\
VADTree~\cite{li2026vadtree}        & \underline{48.81} & \xmark & \cmark \\
EventVAD~\cite{shao2025eventvad}    & 48.54 & \xmark & \cmark \\
LAVAD~\cite{zanella2024harnessing}  & 45.54 & \xmark & \cmark \\
AnomalyRuler~\cite{yang2024follow}  & 39.59 & \xmark & \cmark \\
\bottomrule
\end{tabular}
\caption{Frame-level AUC (\%) on the SENSE-VAD test set by supervision paradigm. Bold = best overall; \underline{underline} = best within paradigm. Real-time = operates at (or near) sensor frame rate without per-window VLM/LLM inference; Explain. = provides natural-language explanations. \cmark/\xmark~= supported/not supported.}
\label{tab:main_results}
  \vspace{-3mm}
\end{table}

\subsection{Analysis by Supervision Paradigm}
\subsubsection{Semi-supervised}
FFP~\cite{liu2018future}, which is trained on nominal clips to predict future frames, reaches only 44.56\% AUC, below chance. This outcome follows from what the method actually measures. Its prediction error is a proxy for motion surprise, whereas SENSE-VAD's anomalies are by design motion-statistically plausible. A child running from a guardian, for instance, moves at the same speed and on the same trajectory as a child running toward one, so the relational signal that separates the two remains invisible to a predictor trained only on motion regularities.
\subsubsection{Weakly-supervised}
VadCLIP (50.80\%) and VERA (50.53\%) remain at chance, indicating that vision-language grounding alone provides no actionable signal. REWARD (37.38\%) and BN-WVAD (37.86\%) fall below chance for a shared reason despite their different scoring criteria. Both flag anomalies by distance from a nominal distribution in the Swin-B feature space, whether by nearest-neighbour distance or BatchNorm-based Mahalanobis distance. Since social anomalies are kinematically ordinary, they lie close to that nominal distribution and are labeled normal, and each method's training loop then propagates the inverted labels. The near-identical AUC confirms that the failure is upstream. Any abnormality criterion defined in this feature space is structurally blind to relational social cues.
\subsubsection{Unsupervised}
The two strongest methods on SENSE-VAD are unsupervised, NP-VAD (57.15\%) and CLAP (56.12\%), yet both remain too low for reliable detection. Each avoids the label-induced bias of the supervised paradigms, but only partially, because their pseudo-labeling still operates over kinematic feature statistics that cannot separate social anomalies from normal behavior. Their marginal advantage therefore reflects the absence of an actively misleading signal rather than any positive relational understanding.
\subsubsection{Training-free foundation-model reasoning}
These methods are among the strongest a priori candidates for SENSE-VAD, since they explicitly target relational and semantic scene understanding; their collective underperformance is therefore the benchmark's most informative result. LAVAD captions frames with BLIP-2 and scores them against a nominal caption index via an LLM, yet the captions encode neither inter-agent relationships nor their absence. AnomalyRuler induces natural-language rules from nominal frames with GPT-4 and applies them at test time with Mistral-7B; its below-chance AUC indicates the induced rules are anti-predictive. A rule such as "a child walks with an adult" does not invert into a reliable detector for the relational violation it describes. VADTree, which chains EfficientGEBD event boundaries with LLaVA-NeXT perception and DeepSeek-R1 reasoning, and EventVAD both remain near chance, consistent with the paradigm-wide pattern.

\subsection{VLM Probing Study}
\label{sec:probe}
 
Benchmarking VAD methods tells us whether SENSE-VAD is hard; a complementary question is \emph{why}. Do AV-oriented vision--language models fail on social anomalies because they cannot see the relevant agents, or because they see them and miss their social significance? To find out, we probe a state-of-the-art VLM with the structured question types of the DriveLM Graph VQA
framework~\cite{drivelm2024} and measure how often it overlooks an anomaly that is directly in view.
We hypothesize that this failure is representational rather than an artifact of prompt design~\cite{hallgarten2024can} ~\cite{xie2025vlmsreadyautonomousdriving}. The following probe is designed to test that hypothesis.
 
\subsubsection{Probe questions.}
We adopt the four DriveLM question types, each targeting a different level of scene reasoning:
\begin{itemize}
    \item \textbf{Perception:} \textit{``What are the important objects in the
          current scene? Describe their type, position, and behaviour.''}
    \item \textbf{Prediction:} \textit{``For each person or vehicle visible,
          predict their future trajectory or intended action in the next few
          seconds.''}
    \item \textbf{Planning:} \textit{``What action should the ego vehicle take
          right now, and why? Consider all objects in the scene.''}
    \item \textbf{Safety:} \textit{``Is there any dangerous or unusual situation
          in this scene that requires immediate attention? Describe it.''}
\end{itemize}
 
\subsubsection{Model and frame sampling.}
We probe Qwen3-VL-4B-Instruct~\cite{bai2025qwen3}, a strong VLM representative of the models used in end-to-end AV reasoning. For each test video we sample five frames from within the annotated anomalous window, guaranteeing that every probed frame actually contains the anomaly, and we apply the same probe to normal test frames as a control.
 
\subsubsection{Response classification.}
Each response is labelled \textsc{Anomaly\_Aware} if it explicitly mentions the anomalous event, \textsc{Driving\_Normal} if it gives a standard driving answer with no acknowledgement of the anomaly, or \textsc{Missed\_Agent} if the anomalous agent is not mentioned at all. 

\subsubsection{Results}
\begin{figure}[t]
\centering
\includegraphics[width=0.95\columnwidth]{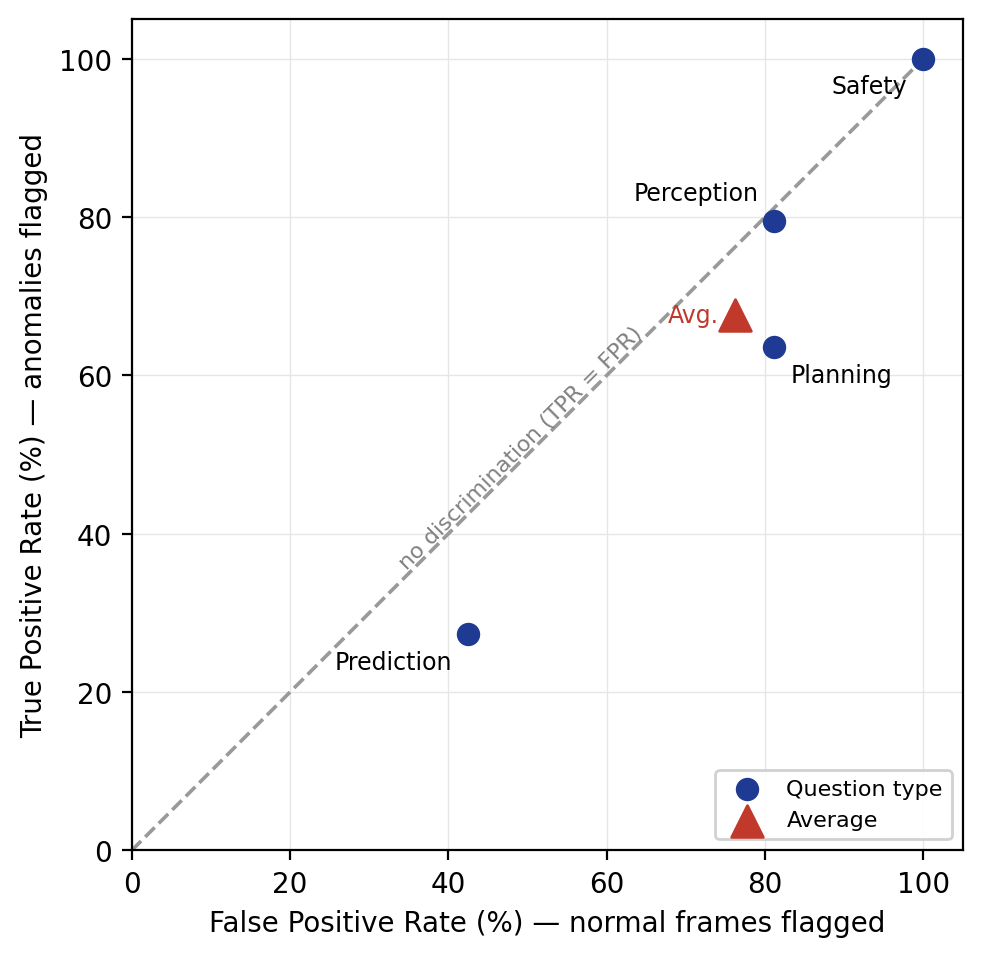}
\caption{VLM anomaly flagging on SENSE-VAD (true positive rate vs. false positive rate) for Qwen3-VL-4B-Instruct. All DriveLM question types lie on or below the no-discrimination diagonal, indicating no meaningful anomaly discrimination. The safety question always predicts an anomaly.}
\label{fig:probe_tprfpr}
  \vspace{-4mm}
\end{figure}

We probe Qwen3-VL-4B-Instruct on 220 anomalous frames (44 videos $\times$ 5) and 80 normal frames (16 videos $\times$ 5), with four questions per frame, for 1{,}200 responses in total. To place the model on a familiar detection axis, we report its responses as a true positive rate (TPR), the fraction of anomalous frames on
which it flags the anomaly, against a false positive rate (FPR), the fraction of normal frames on which it erroneously reports an anomaly. A competent detector would sit toward the top-left of this space (high TPR, low FPR); the diagonal $\text{TPR}=\text{FPR}$ marks the locus of no discrimination, i.e., random guess, where
the model is equally likely to raise an anomaly whether or not one is present. Figure~\ref{fig:probe_tprfpr} plots each DriveLM question type together with the average across question types.
 
The results in Figure~\ref{fig:probe_tprfpr} show that no question type escapes the random guess performance toward the detector region. Three observations make this concrete. \emph{First}, the safety question sits exactly at $(100,100)$: the model answers ``dangerous'' on every frame, anomalous or normal alike. Its perfect TPR is therefore meaningless; paired with an equally perfect FPR, it carries no discriminative information, consistent with the language-prior dominance report~\cite{xie2025vlmsreadyautonomousdriving}. \emph{Second},
perception lies almost on the diagonal at $(81.2, 79.5)$: the model mentions the relevant agents at nearly identical rates on anomalous and normal frames, so its high recall reflects generic scene description rather than anomaly awareness.
\emph{Third}, the planning and prediction points fall \emph{below} the diagonal ($(81.2, 63.6)$ and $(42.5, 27.3)$ respectively): when reasoning about what the ego vehicle should do, or what agents will do next, the model is actually less likely to surface the anomaly on anomalous frames than to raise a false alarm on
normal ones. The anomaly does not propagate into the model's decisions or its forecasts even though the agents themselves are detected.
 
Taken together, the average operating region (TPR $67.6\%$, FPR $76.2\%$) places the model below the line of chance. Its tendency to report an anomaly is statistically unrelated to whether a social anomaly is present. The near-coincidence of perception's TPR and FPR localises the failure precisely, the model sees the
agents but does not register the relational context that makes the scene anomalous. This is a breakdown of social interpretation, not object detection, and because it persists across all four question formulations it cannot be resolved by prompt engineering alone. Closing it requires training on data dedicated to social anomalies, which is what SENSE-VAD can provide.

\section{Conclusion and Future Work}
\label{sec:conclusion}

We introduced SENSE-VAD, the first synthetic, egocentric VAD benchmark for autonomous driving built around socially complex anomalies. Events whose danger lies in inter-agent relationships rather than motion statistics with 15 anomaly classes across six categories and 540,888 per-frame-labeled frames. Its guiding principle, separating social anomaly from motion-statistics anomaly, distinguishes it from existing AV datasets, none of which annotate relational inter-agent structure from the onboard viewpoint. Benchmarking 11
state-of-the-art methods confirms that socially complex anomalies remain a genuine and currently unsolved challenge.

SENSE-VAD shows the potential for several natural extensions, which we group into two directions. The first concerns the scope of the benchmark itself. Because its scenarios are generated through simulation rather than real-world capture, the benchmark can grow without the cost or safety risk of staging new events, and we plan to broaden its coverage of interaction-heavy urban situations, including cyclist–pedestrian conflicts, emergency-vehicle yielding, and multi-agent cascades. We further intend to enrich the annotations with spatial agent localization, natural-language rationales at anomaly onset, and finer-grained severity levels, which together would extend the benchmark from detection alone to localization and explainability.
Additionally, our results indicate a clear objective for method design. Reliable performance on socially complex anomalies will require detectors that explicitly represent inter-agent relational structure, for example through scene-graph or object-centric models, together with training protocols that provide social-context supervision directly rather than relying on zero-shot foundation-model queries. By pairing behaviorally matched normal and anomalous footage, SENSE-VAD is constructed to support this form of training, and we regard it as a basis for AV perception systems that anticipate social anomalies rather than react to them after they occur.

{
    \small
    \bibliographystyle{ieeenat_fullname}
    \bibliography{main}
}
\clearpage
\appendix
\section{Appendix}
\subsection{Datasheet for SENSE-VAD}
\label{app:datasheet}
This datasheet of SENSE-VAD follows the framework of Gebru et al.\footnote{Timnit Gebru, Jamie Morgenstern, Briana Vecchione, Jennifer Wortman Vaughan, Hanna Wallach, Hal Daum´e III, and
Kate Crawford. Datasheets for datasets, 2021}. 

\subsubsection{Motivation}

\textbf{For what purpose was the dataset created?} SENSE-VAD was created to benchmark video anomaly detection (VAD) for autonomous driving on socially complex anomalies — events whose danger arises from inter-agent relationships rather than motion or appearance statistics (e.g., a child diverging from a guardian, a person being carried, police chasing a pedestrian). Existing AV anomaly datasets define anomalies through vehicle kinematics or accident outcomes and do not annotate relational social structure from an egocentric viewpoint; SENSE-VAD was built to fill that gap and to test whether current methods, including VLM/LLM-based ones, can detect such anomalies.

\textbf{Who created the dataset and on behalf of which entity?} Omitted for anonymous review; will be disclosed upon acceptance.

\textbf{Who funded the creation of the dataset?}
The dataset was created using internal resources of the authors' research laboratory. No external or third-party funding was involved.

\subsubsection{Composition}

\textbf{What do the instances represent?} Instances are video clips recorded from an egocentric (AV dashcam) viewpoint, each accompanied by per-frame binary anomaly labels. Most clips are synthetic (CARLA / Unreal Engine); a smaller set of 4 anomalous and 2 normal real-world clips are recorded from a moving car or motorcycle.

\textbf{How many instances are there in total?}
108 clips totaling 540,888 frames (171.2 minutes). Composition: 65 anomaly clips and 43 normal clips. Normal footage additionally includes 28 DAVID sequences (10,767 frames). Real-world clips: 4 anomalous, 2 normal. Train/test split: 48 train clips (262,534 frames) / 60 test clips (278,354 frames).

\textbf{Does the dataset contain all possible instances or is it a sample?}
It is a constructed/sampled benchmark, not an exhaustive population. Each of the 15 anomaly classes is represented by 1–5 clips. Scenarios were authored to instantiate a defined taxonomy rather than to be statistically representative of real-world frequency.

\textbf{What data does each instance consist of?}
Raw video frames (egocentric RGB, 800 × 600, 25–60 FPS) plus a per-frame binary label vector. For multi-interval clips, the per-clip (onset, offset) intervals are provided.

\textbf{Is there a label or target associated with each instance?}
Yes, per-frame binary anomaly labels (normal/anomalous), assigned by the onset protocol in which all frames from the first unambiguous anomalous frame to clip end are labeled anomalous.

\textbf{Is any information missing from individual instances?}
The current release provides temporal (per-frame) labels only. Spatial annotations (bounding boxes) are provided for the Anomaly test set only, and natural-language rationales are noted as future work and are not included.

\textbf{Are there recommended data splits?}
Yes. A train/test split is provided and deliberately weighted toward evaluation. Three classes (Kid Occlusion, Cyclist–Vehicle Collision, Kidnapping) appear only in test, yielding an open-set evaluation with unseen anomaly types.

\textbf{Are there any errors, sources of noise, or redundancies?}
The labeling scheme is intentionally conservative: ambiguous frames are labeled normal to avoid inflating anomaly recall, so some genuinely transitional frames may be under-labeled as anomalous.

\textbf{Is the dataset self-contained or does it rely on external resources?}
It relies on external resources: the DAVID normal sequences (Kaggle) and the CARLA simulator / City Sample assets.

\textbf{Does the dataset contain confidential data?}
No confidential data.

\textbf{Does the dataset contain data that might be offensive, insulting, threatening, or otherwise cause anxiety?}
Yes, by design the taxonomy includes distressing scenarios (e.g., Kidnapping, Bag Theft, Person Being Carried, collisions). However, all such scenarios involving children or coercion are synthetic. 

\textbf{Does the dataset identify any subpopulations / contain data that could identify individuals?}
Synthetic clips depict simulated agents (no real persons). Real-world clips depict consenting adult participants in staged scenarios. Additionally, no bystanders are clearly identifiable.

\textbf{Any other comments?}
All child-involving and coercion scenarios are simulated; text-to-video generators were explicitly rejected in part because they prohibit child characters by content policy.

\subsubsection{Collection Process}

\textbf{How was the data acquired?} Synthetic clips were generated as scripted episodes in CARLA (versions 0.9.12, 0.9.15, and 0.10.0 on UE5), with controlled agent placement, camera configuration, weather, and timing. Real-world clips were recorded from a moving car or motorcycle.

\textbf{What mechanisms or procedures were used to collect the data?}
Parameterized generation scripts (released) for the synthetic portion; physical dashcam-style recording for the real-world portion. Camera parameters were held constant across scenarios for comparability.

\textbf{Who was involved in the data collection process?}
All data collection, synthetic scene authoring and generation in CARLA, and recording of the real-world clips, was carried out by the authors and author's labmate. No external annotators, crowdworkers, or third-party contractors were involved. Author identities are withheld to preserve anonymity during double-blind review.

\textbf{Over what timeframe was the data collected?
}
The dataset was generated and recorded over a period of approximately 18 months, from January 2025 to June 2026.

\textbf{Were any ethical review processes conducted (e.g., by an institutional review board)?}
No formal institutional review was conducted. The real-world clips involved only adult volunteers who consented to be recorded performing staged, non-hazardous actions, and no personally identifying or sensitive information is released. All other scenarios, including ones involving children or coercion are entirely synthetic and depict no real persons.

\textbf{Did you collect data from individuals directly, and were they notified / did they consent?}
Real-world clips depict staged adult-only anomalies performed with consent.

\subsubsection{Preprocessing / Cleaning / Labeling}

\textbf{Was any preprocessing/cleaning/labeling done?}
Yes. Per-frame binary labels were assigned via the onset protocol (§2). For clips with multiple anomaly intervals, the (onset, offset) pairs are unioned into a single label vector. For evaluation, each method's frame-level scores are linearly interpolated to the clip's frame count before AUC computation.

\textbf{Was the raw data saved in addition to the preprocessed data?}
Yes, everything from the raw data was saved, including the raw footage, and the scene generation code.

\textbf{Is the software used to preprocess/label the data available?}
The parameterized scene-generation scripts are released. Frame-level annotation was performed using CVAT (Computer Vision Annotation Tool), an open-source third-party tool that is publicly available and not redistributed here. The resulting per-frame annotations are released in json together with the conversion scripts used to derive the final per-frame binary label vectors from the CVAT exports.

\subsubsection{Uses}

\textbf{Has the dataset been used for any tasks already?}
Yes, frame-level VAD benchmarking of 11 state-of-the-art methods across four supervision paradigms, plus a VLM probing study (Qwen3-VL-4B-Instruct with DriveLM question types).

\textbf{Is there a repository linking papers/systems that use the dataset?}
Not at this time, as the dataset has not yet been publicly released, no external papers or systems currently use it. Upon publication, the dataset will be made available via the lab's website, and a list of papers and systems that use it will be maintained at that location.

\textbf{What (other) tasks could the dataset be used for?}
Open-set anomaly detection, sim-to-real transfer evaluation, scene-graph / relation-aware perception research, and (with planned extensions) anomaly localization and explanation.

\textbf{Is there anything about the composition or collection that might impact future uses?}
Anomaly frequencies are not naturalistic (constructed benchmark), the synthetic portion carries simulator-specific visual characteristics, and the resolution (800 × 600) and class coverage are limited. Users should not treat per-class clip counts as real-world priors.

\textbf{Are there tasks for which the dataset should not be used?}
It should not be used to train production safety systems as a sole data source, nor as evidence of real-world anomaly base rates.

\subsubsection{Distribution}

\textbf{How will the dataset be distributed?
}Via Zenodo (DOI record cited in the paper).

\textbf{When will it be distributed?}
The dataset is already distributed. An anonymized version has been made publicly available for review via Zenodo, and the full, de-anonymized release — including author and institutional details — will follow upon publication.

\textbf{Will it be distributed under a license / terms of use?}
Yes. The SENSE-VAD annotations, data, and the scene-generation and labeling code under Apache-2.0. The synthetic video frames are distributed as rendered, non-interactive media.

\textbf{Have any third parties imposed IP-based or other restrictions?}
Yes. Three third-party components carry their own terms, which we do not re-license:
\begin{itemize}
    \item CARLA: code is under the MIT License and assets under CC-BY; use requires attribution to the CARLA project.
    \item Unreal Engine 5 City Sample — designated UE-Only Content under the Unreal Engine EULA, usable only in conjunction with the engine. Accordingly, we redistribute only rendered, non-interactive video output produced with the engine; the underlying City Sample assets are not included in our release and may not be extracted or reused outside Unreal Engine.
    \item DAVID dataset — licensed under CC BY 4.0, requiring attribution to the original author and an indication of changes.
\end{itemize}
All other content in SENSE-VAD is our own and is released under Apache-2.0. These third-party terms constrain redistribution of the corresponding components but do not restrict use of the SENSE-VAD annotations or generation code.

\textbf{Do any export controls or regulatory restrictions apply?}
None that we are aware of. The dataset consists of simulated and consensually recorded video for academic research and contains no controlled technical data, personal data, or other content subject to export-control or regulatory restrictions.

\subsubsection{Maintenance}

\textbf{Who will be supporting/hosting/maintaining the dataset?
}The dataset is hosted on Zenodo, which provides long-term archival storage and a persistent DOI. It will be maintained by the authors. Specific maintainer names and institutional affiliation are withheld to preserve anonymity during double-blind review and will be provided in the camera-ready version upon acceptance.

\textbf{Is there an erratum?}
No. This is the initial release of the dataset and no errata exist at this time. Any future corrections will be documented in the change-log of the dataset record and reflected in a new versioned release on Zenodo.

\textbf{Will the dataset be updated?}
Yes — planned extensions include broader interaction-heavy scenarios (cyclist–pedestrian conflicts, emergency-vehicle yielding, multi-agent cascades), spatial localization annotations, natural-language rationales, and finer-grained severity levels.

\textbf{If others want to extend/build on the dataset, is there a mechanism?}
Yes, the released parameterized generation scripts allow authoring new scenarios.

\subsection{Reactive vs.\ Anticipatory Avoidance: A Qualitative Example}
\label{app:reactive_example}
 
A publicly shared clip of a commercial autonomous vehicle illustrates the distinction our benchmark is built around.\footnote{Operator demonstration video, shared December 2024: \url{https://x.com/dmitri_dolgov/status/1868778679868047545}.
Accessed 2026-06-25. Our description reflects the publicly visible footage; we make no claim about the vehicle's internal state or sensor data.} In the clip, a scooter rider ahead of the ego vehicle drifts toward the curb and loses balance; as the rider falls into the roadway, the vehicle performs a rapid lateral maneuver into the adjacent lane and avoids contact.
 
The avoidance is effective, but it is fundamentally \emph{reactive}: the
maneuver is triggered by the fall itself, a sudden change in the obstacle's position. The cues that preceded the fall, however, were \emph{social and physical} in nature and, in principle, anticipatory: a rider drifting laterally toward the curb at reduced stability is a recognizable precursor to loss of control, of the kind an attentive human driver may read as a reason to slow and increase following distance \emph{before} any obstacle enters the lane. A system
that models this relational and behavioral context could downgrade its speed in advance, converting a last-moment evasive swerve into a gradual, lower-risk deceleration.
 
This is precisely the capability gap SENSE-VAD is designed to measure. Motion- and appearance-based detection responds to the anomaly once it has materialized as a kinematic event; anticipating it requires interpreting the behavioral trajectory of an agent before the event occurs. Several SENSE-VAD classes encode exactly such precursors, an unstable cyclist (\emph{Cyclist Hazard}), a person losing motor control (\emph{Abnormal Person Behavior}), a child diverging from a guardian, where the safety-relevant signal is available well before any
collision-like motion appears.

\subsection{Benchmarked Baseline Details}
\label{sec:baseline_details}

\newcolumntype{Y}{>{\raggedright\arraybackslash}X}
\begin{table*}[t]
\centering
\footnotesize
\setlength{\tabcolsep}{4pt}
\renewcommand{\arraystretch}{1.15}
\begin{tabularx}{\textwidth}{@{}l l Y@{}}
\toprule
\textbf{Method} & \textbf{Backbone / foundation model(s)} & \textbf{Mechanism and adaptation to SENSE-VAD} \\
\midrule
\multicolumn{3}{@{}l}{\textit{Semi-supervised --- normal-only training}}\\
\cmidrule{1-3}
FFP & U-Net + RAFT-Small flow & Future-frame prediction trained on nominal clips; PSNR of prediction error as score. RAFT replaces FlowNet2 to drop legacy CUDA. \\
\addlinespace[2pt]
\multicolumn{3}{@{}l}{\textit{Weakly-supervised --- video-level MIL}}\\
\cmidrule{1-3}
VadCLIP & CLIP ViT-B/16 & Frozen CLIP features with learnable prompts under MIL; anomaly head retrained on SENSE-VAD weak labels. \\
VERA & Qwen2-VL-7B (LoRA) & Guided five-question VQA; LoRA fine-tuned on the training split; snippet scores propagated to frames. \\
REWARD & Video Swin-B & End-to-end metric learning with kNN pseudo-labels for real-time wVAD; reimplemented with a Swin-B extractor. \\
BN-WVAD & Video Swin-B & BatchNorm running-mean Mahalanobis distance (DFM) as abnormality criterion; reimplemented on Swin-B features. \\
\addlinespace[2pt]
\multicolumn{3}{@{}l}{\textit{Unsupervised --- no labels}}\\
\cmidrule{1-3}
NP-VAD & Video Swin-B & Temporal-boundary normality prior propagated over a similarity graph to derive pseudo-labels; Swin-B features. \\
CLAP & MLP on Swin-B feats. & GMM clustering on feature-statistic descriptors yields pseudo-labels refined by hypothesis testing; federated stage reduced to local training. \\
\addlinespace[2pt]
\multicolumn{3}{@{}l}{\textit{Training-free --- foundation-model reasoning}}\\
\cmidrule{1-3}
LAVAD & BLIP-2 + LLaMA-2-13B & Frame captions scored by an LLM via FAISS caption/summary indices; nominal index built from training clips. \\
AnomalyRuler & CogVLM + GPT-4 + Mistral-7B & GPT-4 induces a natural-language rule set from nominal frames; Mistral applies the rules zero-shot at test time. \\
VADTree & EfficientGEBD + LLaVA-NeXT + DeepSeek-R1 & Granularity-aware event tree with VLM node perception and LLM reasoning; applied directly to test clips. \\
EventVAD & CLIP ViT-B/16 + RAFT-Large + VideoLLaMA2.1-7B & Dynamic-graph event segmentation over CLIP and flow features; each segment scored by a VLM via chain-of-thought; applied directly to test clips. \\
\bottomrule
\end{tabularx}
\caption{Benchmarked baselines, grouped by supervision paradigm.}
\label{tab:baselines}
\end{table*}

Table~\ref{tab:baselines} lists the eleven baselines and their backbones; here we report the adaptations and settings not covered in Sec~\ref{sec:baselines}. Method selection spans all four supervision paradigms and over-samples the semantic VLM/LLM family, the strongest \textit{a priori} candidates for relational anomalies. Where an original extractor was unavailable (REWARD, BN-WVAD,
NP-VAD, CLAP) we substitute a shared Video Swin-B to control for backbone confounds; features are computed on non-overlapping 16-frame windows at the native $800\times600$ resolution, and every method's raw scores are min--max normalized per clip before the interpolation described in Sec~\ref{sec:eval_protocol}. For the
open-set classes (Kid Occlusion, Cyclist--Vehicle Collision, Kidnapping), no clip of those classes appears in any nominal reference, preserving the unseen condition across all paradigms.

\paragraph{Semi-supervised.}
FFP trains on the nominal split only (50 epochs, Adam optimizer with $2\times10^{-4}$ LR) with RAFT-Small replacing FlowNet2 to remove a legacy-CUDA dependency; the prediction objective and PSNR scoring are unchanged.

\paragraph{Weakly-supervised.}
VadCLIP and VERA update only their anomaly head and LoRA adapter (rank 8, 50 epochs) on the training split, with the CLIP / Qwen2-VL backbones frozen. REWARD and BN-WVAD estimate their nominal statistics (kNN reference set and BatchNorm running moments, respectively) from the training normals only, so no anomalous frame informs the criterion.

\paragraph{Unsupervised.}
NP-VAD and CLAP receive no anomaly labels. We reduce CLAP's federated stage to a single local-training run to match our single-node protocol; remaining hyperparameters follow the original implementations.

\paragraph{Training-free.}
These methods require no parameter updates and run directly on the test clips; any nominal reference is built from the training split. LAVAD's FAISS index is populated from approximately $5{,}500$ nominal captions/summaries. AnomalyRuler is trained on nominal training frames, uses CogVLM descriptions and GPT-4 to induce a domain-adapted rule set, then applies the rules zero-shot via Mistral-7B-Instruct-v0.2 with RAM2 semantic tags per test frame; scores are post-processed with a first-pass EMA ($\alpha{=}0.33$), 21-frame majority voting, and a second EMA ($\alpha{=}\bar{\mu}$).

\end{document}